\documentclass[10pt,twocolumn,letterpaper]{article}

\usepackage{cvpr}
\usepackage{times}
\usepackage{epsfig}
\usepackage{graphicx}
\usepackage{amsmath}
\usepackage{amssymb}
\usepackage{booktabs}

\graphicspath{{./Figures/}}
\usepackage{array}
\usepackage{caption}
\usepackage{subcaption}
\usepackage{multirow}

\newcommand{\ig}[2]{\includegraphics[width=#1]{#2} } 
\newlength{\sm}
\newcommand{\rt}[2]{\parbox[t]{2mm}{\multirow{#2}{*}{\rotatebox[origin=c]{90}{\textbf{#1}}}}}

\usepackage[pagebackref=true,breaklinks=true,letterpaper=true,colorlinks,bookmarks=false]{hyperref}

\cvprfinalcopy 

\def\cvprPaperID{3314} 

\ifcvprfinal\fi
\begin{document}

\title{Visual Saliency Prediction Using a Mixture of Deep Neural Networks}

\author{Samuel Dodge and Lina Karam\\
Arizona State University\\
{\tt\small \{sfdodge,karam\}@asu.edu}
}

\maketitle

\begin{abstract}
  Visual saliency models have recently begun to incorporate deep learning to achieve predictive capacity much greater than previous unsupervised methods. However, most existing models predict saliency using local mechanisms limited to the receptive field of the network. We propose a model that incorporates global scene semantic information in addition to local information gathered by a convolutional neural network. Our model is formulated as a mixture of experts. Each expert network is trained to predict saliency for a set of closely related images. The final saliency map is computed as a weighted mixture of the expert networks' output, with weights determined by a separate gating network. This gating network is guided by global scene information to predict weights. The expert networks and the gating network are trained simultaneously in an end-to-end manner. We show that our mixture formulation leads to improvement in performance over an otherwise identical non-mixture model that does not incorporate global scene information.
  
\end{abstract}

\section{Introduction}

Visual attention enables the human visual system to efficiently process the flood of visual information entering the retina. This mechanism enables the visual system to focus resources on the most relevant locations in the scene. The tendency of a particular region in a scene to receive attentional focus can be represented by the visual saliency of that region.

The study of computational models of this mechanism may increase our understanding of the underlying biological mechanisms, but also can lead to applications in computer vision. In these applications, saliency can be used for the same purpose as in the human visual system: to focus resources on the key parts of the scene for more efficient processing of the visual world. This has found application in object detection \cite{walther}, scene recognition \cite{borji-scene}, and robotic navigation~\cite{frintrop-slam}, among others.

Visual attention is often said to be composed of top-down and bottom-up components \cite{bot-top}. Existing deep neural network based models of visual saliency can be said to encode both factors: bottom-up information can be manifested in the outputs from filters in early layers of a convolutional neural network, and top-down information (such as the location of faces) can manifest itself in later layers. However, top-down information can go beyond recognition of familiar objects to include prior experience \cite{sun}, scene semantics and context \cite{henderson-real} and task information~\cite{yarbus}.

Task and prior experience are difficult to model, so we focus our attention on scene context. Several studies show the effect of scene context on saliency in the form of contextual cueing \cite{henderson-real, chun, kunar}. In these experiments, the time to visually locate a target is reduced when the target appears in a previously seen arrangement of distractors \cite{chun}, in consistent global colors \cite{kunar}, or more generally in natural images \cite{henderson-real}. Contextual cueing tells us that humans adapt and learn contextual information to find more optimal visual search strategies. Similarly, we propose a computational model of attention that can adapt to different contextual information.

We introduce a saliency model that learns a measure of global scene contextual information. In our model, global scene information corresponds to different categories of image stimuli (e.g., natural images, fractal patterns, etc.). We posit that these categories are varied enough to induce different saliency mechanisms. Therefore we learn a different prediction for each category. These predicted saliency maps are generated using an efficient tree structure to share common bottom-up features, and diverge in later layers to compute semantic higher-level features. A context guided gating network decides, given an unknown image, weights to give to each prediction. The context gating network and the category-specific saliency predictions are implemented as convolutional neural networks that can be jointly trained. We build our model on a recent deep-learning based model \cite{ml-net}, and show that the additional global scene information leads to greatly increased performance.

\vspace{-12pt}
\paragraph{Related Work}

Early models of visual saliency use biological analogs or concepts from information theory. We can think of these methods as ``unsupervised'' because they do not use any training data. One of the first computational models of visual attention was the Itti model \cite{itti}. This model is based on biological principles of center-surround and feature integration theory. The Itti model was extended by using random walks on a graph structure in the GBVS model~\cite{gbvs}. Another class of models use information theoretic approaches to quantify salient regions of the image \cite{aim}. Saliency can also be related to the frequency response of an image \cite{HouZhang}. Finally, the best performing of the unsupervised models is based on boolean map theory \cite{bms}. A survey of many of these unsupervised methods can be found in \cite{borji-survey}.

Supervised models provide an alternative to the biological or information theoretic models. Kienzle \etal propose a model that uses a SVM to learn which image patches contribute to saliency~\cite{kienzle}. The Judd model \cite{judd} learns a linear combination of many hand-chosen low-level features (center surround, filter responses, etc.) and a few high-level features (face detectors, etc.) to predict saliency. Borji \etal~\cite{borji-boost} extend this further by using a boosting based model with additional features.


The limitation of the aforementioned machine learning approaches is that they are largely dependent on the features used for learning. This can be termed shallow learning because the features are pre-defined. By contrast, deep learning approaches are able to learn rich hierarchies of features from the original pixel data. This type of learning has been made possible by larger scale eye-tracking datasets \cite{salicon-db, cat2000}. Pan \etal \cite{junting} train separate deep and shallow architectures from scratch to predict saliency and show that the deep architecture achieves better performance compared to the shallow architecture. It has also been shown that using deep networks pre-trained on image classification tasks can prove useful for saliency detection. The most commonly used pre-trained networks are the VGG networks~\cite{vgg}. DeepGaze~\cite{deep-gaze1, deep-gaze2} show that the deep features from VGG networks can be used without modification to predict saliency. Other works fine-tune the parameters of the VGG network. The Salicon model \cite{salicon} incorporates a multi-scale approach in addition to fine-tuning. Deepfix~\cite{deepfix} improves performance by adding inception modules \cite{inception}, using dilated convolutions, and explicitly modeling the center bias. Vig \etal~\cite{vig} also fine-tune a VGG-like model but compare the performance when using different cost functions in the training. ML-Net~\cite{ml-net} draws information from the last 3 convolutional layers of the VGG network, instead of only the last layer as in the other networks. 

It could be argued that the existing deep network approaches are capable of modeling scene context and top down information. Top-down concepts such as faces are easily detected and labeled as salient by existing models. Additionally, some degree of local scene context can be modeled. However, the deep networks are limited by the receptive field of an output pixel in relation to input pixels. This receptive field often does not cover the entire input, so the global characteristics of the scene are not modeled. To achieve such large receptive fields, a network could have convolutions with very large input strides, but this may make modeling local saliency more difficult. We propose an alternative solution to incorporating global context by using a set of expert networks and a gating network to weight the experts.

The most closely related work to ours is the iSEEL \cite{iseel} model, which was made available shortly before the submission of this manuscript. For a given image, the model finds similar images in a scene-bank using ``gist'' features and ``classeme'' features from the last layer of the VGG16 network. For each similar image, a separate trained Extreme Learning Machine (ELM) predicts saliency based on VGG16 convolutional features. The final saliency map is computed by summing the outputs of the ELMs. Our model achieves better performance because each of the experts in our model is trained on a collection of images instead of a single image as in the iSEEL model. This allows the experts to generalize better to scene characteristics that are common across many images. Furthermore our gating network is trained along with expert networks in an end-to-end fashion, compared with the fixed features and euclidean distance used in iSEEL to retrieve similar images. In Section \ref{sec:results}, we show that our model achieves better performance on the CAT2000 dataset.

Many of the aforementioned works incorporate information from bottom-up local sources. In addition to bottom-up information, several models have shown that a notion of global scene ``gist'' can be used to help predict saliency~\cite{torralba2006, itti-peters}. In Torralba \etal \cite{torralba2006}, global gist features are used to modulate low level features to compute a task-based attention map that searches for a target object. For example in a city scene, attention can be modulated to look for pedestrians. Peters and Itti \cite{itti-peters} use gist features to learn task based attention for the task of playing video games.


\begin{figure*}[htb!]
  \centering
  \includegraphics[width=0.71\textwidth, natwidth=1989,natheight=1186]{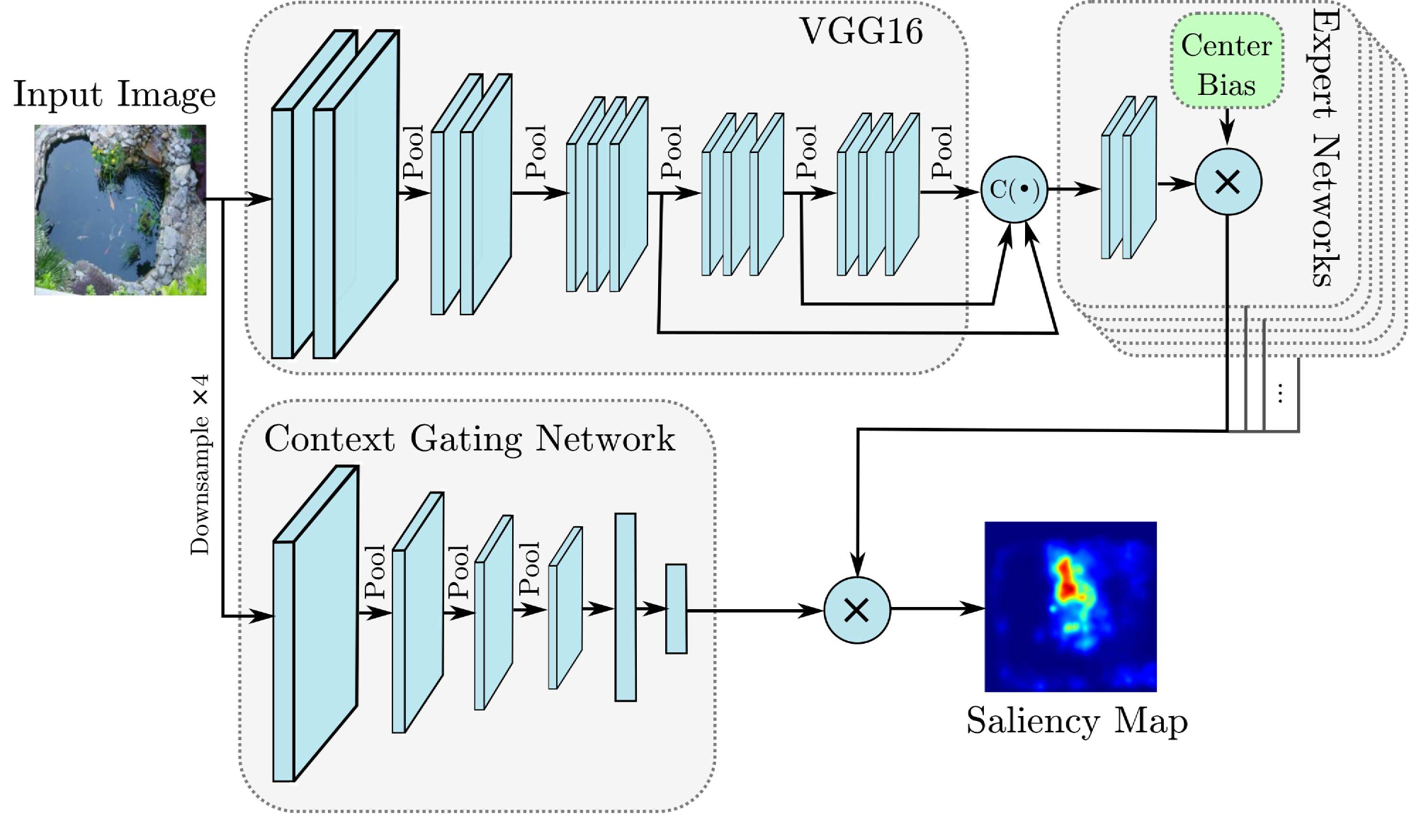}
   \caption{\textbf{Mixture model for predicting visual saliency}. The image is first passed through the convolutional layers of the VGG16 network. Then the responses from the last 3 stages are concatenated. The output of this concatenation is fed to 20 category specific expert networks. A gating network determines weights for the outputs of each of the 20 experts, and the final saliency map is a weighted sum of the expert saliency maps. The parameters for the layers can be found in Table~\ref{table:params}.}
\label{fig:mixture}
\end{figure*}

Our work is related to the gist-based models that incorporate scene context, however we make several improvements that lead to much greater performance. First, instead of using fixed gist features, we learn global scene information from labeled training data. The labels provide hard classes, but we are able to learn a soft weighting function. This weighting function encodes information similar to the ``dark-knowledge'' in supervised neural networks~\cite{hinton-distillation}. Secondly, we utilize recent advances in deep learning to learn a local saliency predictor. We combine the global scene information with the more local saliency information using a mixture of experts formulation \cite{hinton-mixture}. Our model achieves better performance compared with a similar model that does not utilize the global scene information, and outperforms several other deep-learning based models. 


\section{Proposed Approach}


The human visual system is faced with processing many varied types of inputs that are not limited to natural images. From experience, the visual system can learn optimal strategies for images with similar contextual characteristics. Similarly, it is difficult for a single computational model to correctly predict saliency with varied input types. In machine learning, this is known as the interference problem, where a single model can be conflicted by vastly different inputs. We propose that saliency can be modeled as a mixture of experts \cite{hinton-mixture}, where the experts are specialists at predicting some subset of possible images. In a typical mixture of experts model, a separate network termed a gating network is used to weight the experts for an unknown input.

The mixture of experts works best when there is a large amount of data available, so that the experts can adapt to the data. Unfortunately, for visual saliency there are only few large-scale eye tracking datasets. Consequently, a single expert will likely over-fit the subset of the data. Instead of using a ``hard'' mixture of experts model where the entire weight is given to a single expert, we use a soft mixture. In this case, each of the experts is given some weight, with larger weights given to classes that the sample is more likely to be associated with. Over-fitting is reduced because each expert sees related images in addition to the subset of images.

We can express our saliency computation at location $x_i$ as the probability of a binary variable $s_i$, where $s_i=1$ denotes a salient location $x_i$, and a value of $0$ denotes a non-salient location.  $s_i$ is conditioned on the value of the image pixels $\mathbf{I}$ and the location $x_i$. We express the mixture model as:
\begin{equation}
  p(s_i | \mathbf{I}, x_i) = \sum_k p(s_i | \mathbf{I}, x_i, C_k) p(C_k | \mathbf{I}, x_i)
\end{equation}
\noindent where $p(s_i| \mathbf{I}, x_i)$ is the probability of saliency given location $x_i$ and the image $\mathbf{I}$, and $p(C_k | \mathbf{I}, x_i)$ is the conditional probability of the image belonging to class $C_k$ given location $x_i$. Using Bayes' rule we obtain:
\begin{equation}
p(s_i | \mathbf{I}, x_i) = \sum_k \frac{p(\mathbf{I} | s_i, x_i, C_k)}{p(\mathbf{I} | x_i, C_k)} p(s_i | x_i,C_k) p(C_k | \mathbf{I}, x_i)
\end{equation}

We assume that the pixel values $\mathbf{I}$ and the location $x_i$ are independent. Similarly, we assume that the class $C_k$ is independent of the locations $x_i$. With these assumptions, we get:
\begin{equation}
  p(s_i | \mathbf{I}, x_i) = \sum_k \frac{p(\mathbf{I} | s_i, C_k)}{p(\mathbf{I} | C_k)} p(s_i | x_i, C_k) p(C_k | \mathbf{I})
\end{equation}
Using Bayes' rule again we can further simplify the expression:

\begin{eqnarray}
    p(s_i | \mathbf{I}, x_i) \! \! \! \! \! &=& \! \! \! \! \! \sum_k \frac{p(s_i | \mathbf{I}, C_k) p(\mathbf{I} | C_k)}{p(s_i | C_k) p(\mathbf{I} | C_k)} p(s | x_i,C_k) p(C_k | \mathbf{I})  \nonumber \\
    &=& \! \! \! \! \! \sum_k \frac{p(s_i | \mathbf{I}, C_k)}{p(s_i | C_k)} p(s_i | x_i,C_k) p(C_k | \mathbf{I}) 
\end{eqnarray}

This simplifies to:
\begin{equation}
  p(s_i | \mathbf{I}, x_i)  =  \frac{1}{p(s_i)} \sum_k p(s_i | \mathbf{I}, C_k) p(s_i | x_i,C_k) p(C_k | \mathbf{I})
\end{equation}
We ignore the term $p(s_i)$ which describes the prior probability of saliency $s_i$. This leads us to the final formulation for our model:
\begin{equation}
    p(s_i | \mathbf{I}, x_i) \propto \sum_k p(s_i | \mathbf{I}, C_k) p(s_i | x_i,C_k) p(C_k | \mathbf{I})
  \label{eq:prob}
\vspace{-8pt}
\end{equation}
From Eq.~\ref{eq:prob} we see that the probability decomposes into three terms. $p(s_i | \mathbf{I}, C_k)$ is the expert prediction of a saliency map given the category of the image. $p(s_i | x_i,C_k)$ is a modulation term that depends on the spatial location in the image and the category. This term is used to model the center bias tendency of eye tracking data. Finally, $p(C_k | \mathbf{I})$ models the probability that the image belongs to a particular class.

\vspace{-12pt}
\paragraph{Expert networks} First we discuss the computation of $p(s_i | \mathbf{I}, C_k)$, the probability of saliency given an image and its class. To compute this probability, we adopt the deep neural network structure from the ML-Net model~\cite{ml-net}. The first stage of this network consists of the convolutional section from the VGG16 model for image classification~\cite{vgg}. The outputs of the last three stages are concatenated together and followed by two additional convolutional layers. The output of the final layer is the saliency map.

The canonical mixture of experts approach would have a separate network for each class $k$ \cite{hinton-mixture}. However, because the base VGG16 model already uses a significant amount of memory and computational resources, constructing a full mixture of experts model is not practical. Instead, we note that the early layers of deep neural networks often encode similar features~\cite{lenc}. Similarly in saliency, bottom-up components such as center-surround differences do not vary with scene context. These bottom-up components can be represented in the early layers of the network. In later layers, higher order concepts influence saliency. These higher order concepts can vary with global context. Therefore, we share early layers across experts and allow the later layers to adapt to the characteristics of the expert's domain. The VGG16 convolutional layers are shared, and the additional two convolutional layers are separate for each expert. This structure can be seen as a form of a TreeNet~\cite{tree-net}. However, instead of merely averaging the ``branches'' of the network, we use the mixture of experts formulation to give a different weight to the branches depending on the input. 

\vspace{-12pt}
\paragraph{Gating network}
\begin{figure*}
\setlength{\tabcolsep}{1pt}
\renewcommand{\arraystretch}{0.8}
  \centering
  \small{
   \begin{tabular}{cccccccc}
    \multicolumn{2}{l}{Action} &
    \multicolumn{2}{l}{Art} &
    \multicolumn{2}{l}{Cartoon} &
    \multicolumn{2}{l}{Indoor} \\
     \ig{60pt}{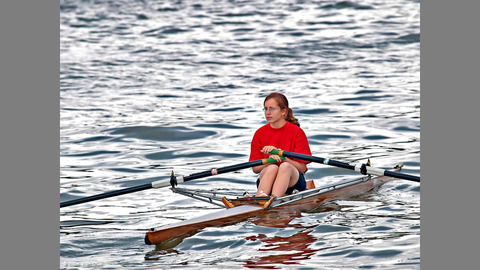} &
     \ig{60pt}{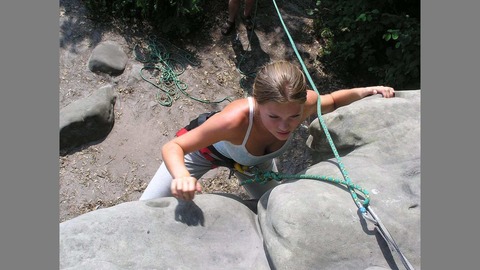} &
     \ig{60pt}{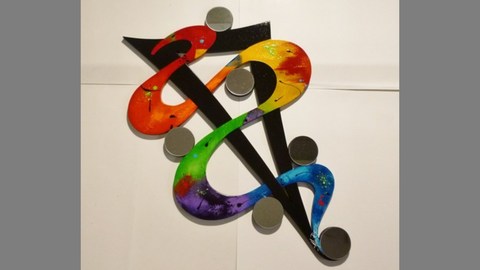} &
     \ig{60pt}{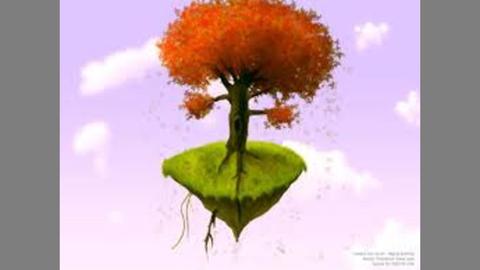} &
     \ig{60pt}{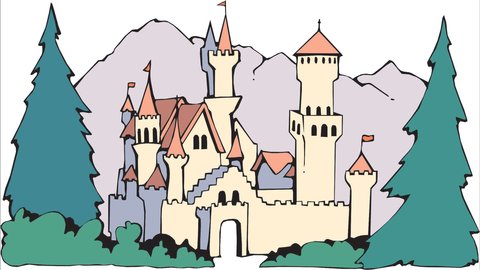} &
     \ig{60pt}{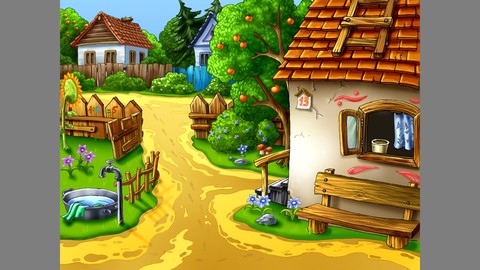} &
     \ig{60pt}{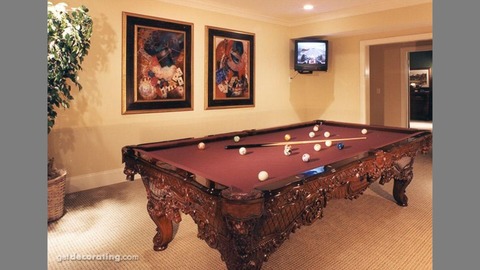} &
     \ig{60pt}{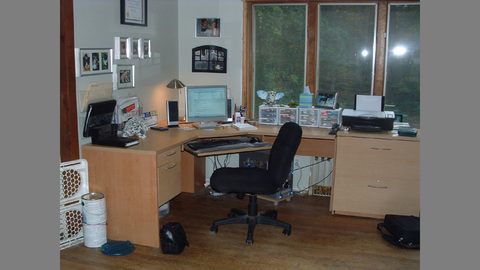} \\
    \multicolumn{2}{l}{Jumbled} &
    \multicolumn{2}{l}{LowResolution} &
    \multicolumn{2}{l}{Object} &
    \multicolumn{2}{l}{OutdoorNatural} \\
     \ig{60pt}{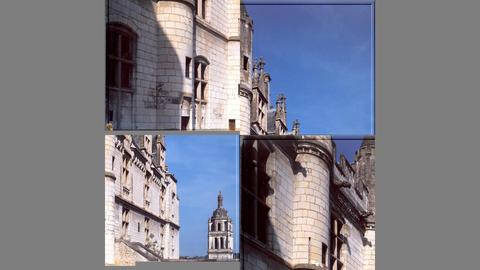} &
     \ig{60pt}{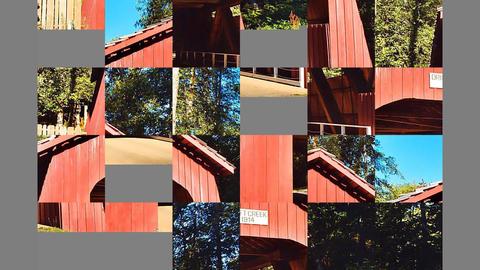} &
     \ig{60pt}{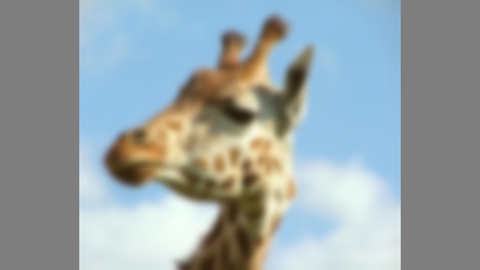} &
     \ig{60pt}{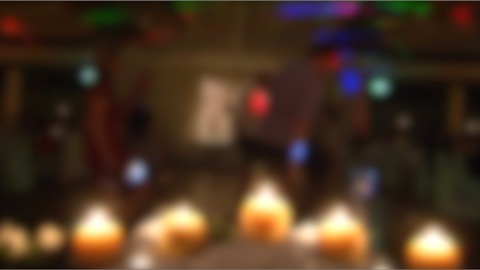} &
     \ig{60pt}{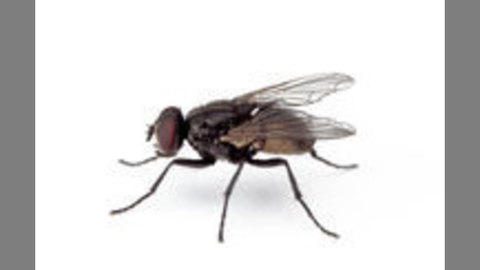} &
     \ig{60pt}{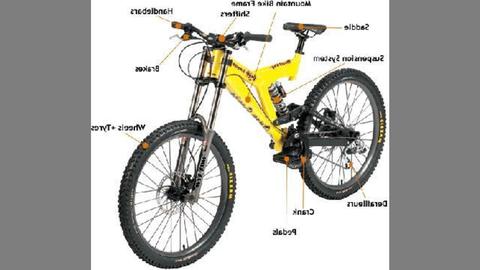} &
     \ig{60pt}{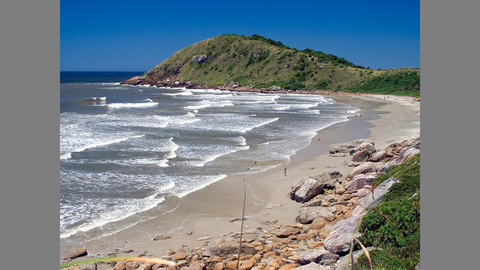} &
     \ig{60pt}{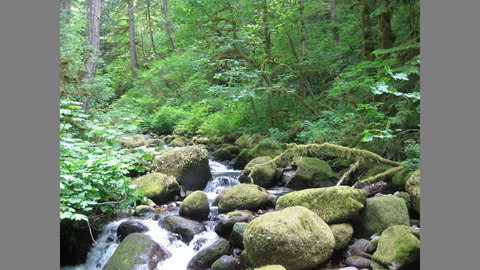} \\
    \multicolumn{2}{l}{Random} &
    \multicolumn{2}{l}{Sketch} &
    \multicolumn{2}{l}{Affective} &
    \multicolumn{2}{l}{BlackWhite} \\
     \ig{60pt}{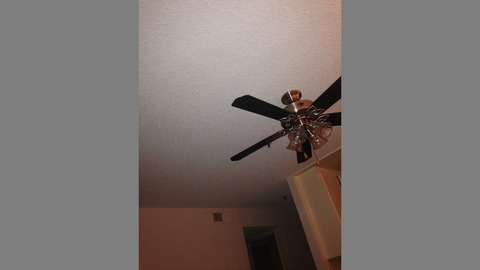} &
     \ig{60pt}{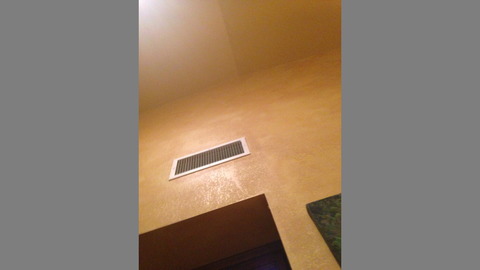} &
     \ig{60pt}{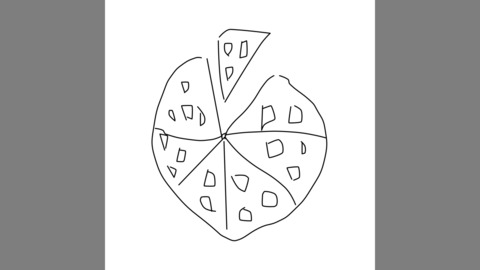} &
     \ig{60pt}{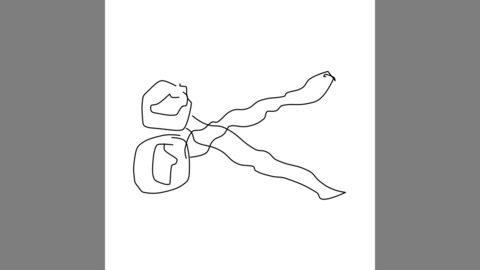} &
     \ig{60pt}{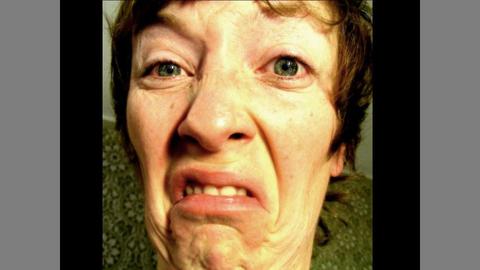} &
     \ig{60pt}{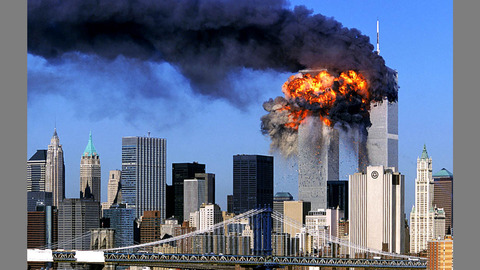} &
     \ig{60pt}{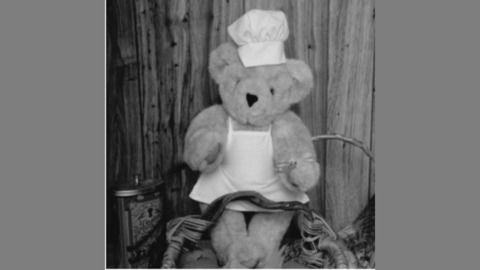} &
     \ig{60pt}{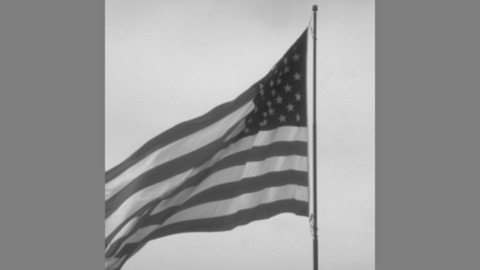} \\
    \multicolumn{2}{l}{Fractal} &
    \multicolumn{2}{l}{Inverted} &
    \multicolumn{2}{l}{LineDrawing} &
    \multicolumn{2}{l}{Noisy} \\
     \ig{60pt}{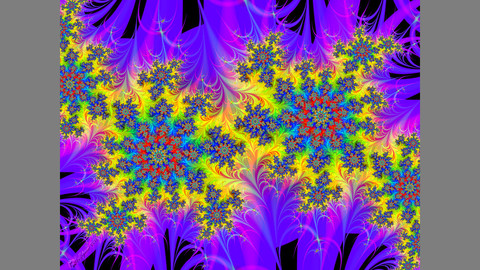} &
     \ig{60pt}{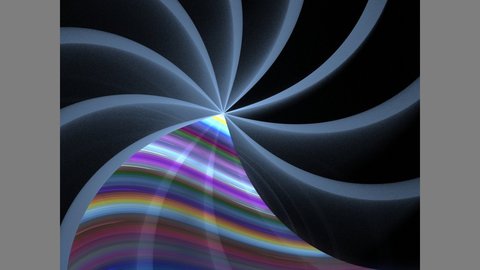} &
     \ig{60pt}{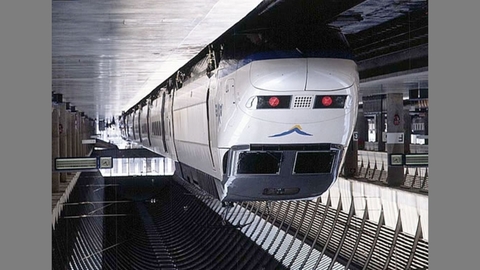} &
     \ig{60pt}{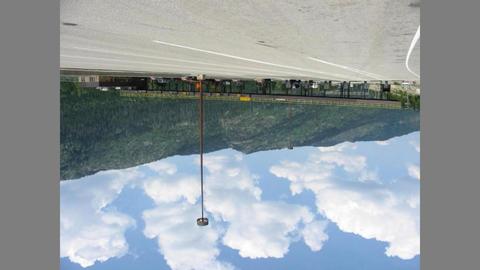} &
     \ig{60pt}{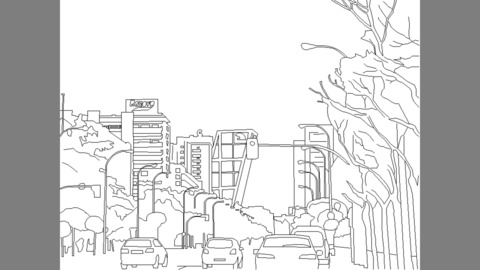} &
     \ig{60pt}{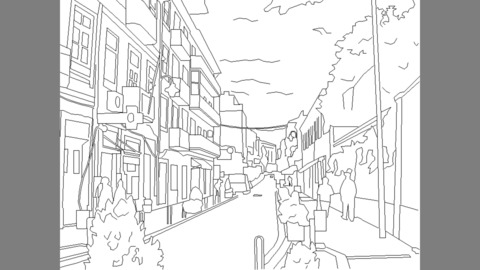} &
     \ig{60pt}{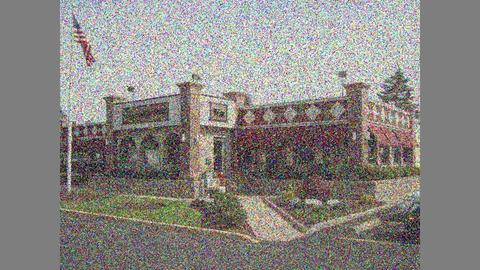} &
     \ig{60pt}{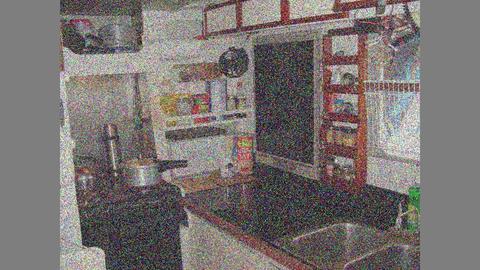} \\
    \multicolumn{2}{l}{OutdoorManMade} &
    \multicolumn{2}{l}{Pattern} &
    \multicolumn{2}{l}{Satelite} &
    \multicolumn{2}{l}{Social} \\
     \ig{60pt}{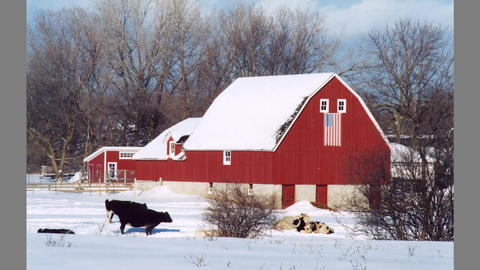} &
     \ig{60pt}{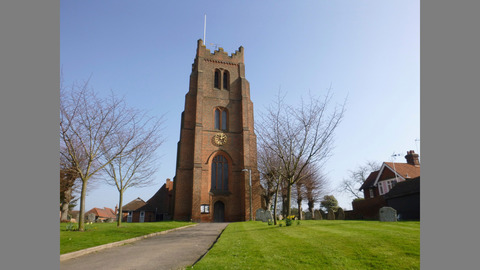} &
     \ig{60pt}{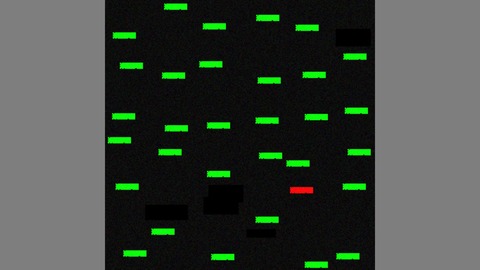} &
     \ig{60pt}{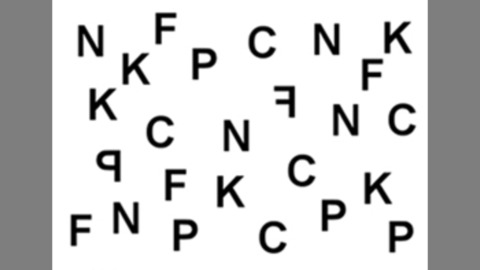} &
     \ig{60pt}{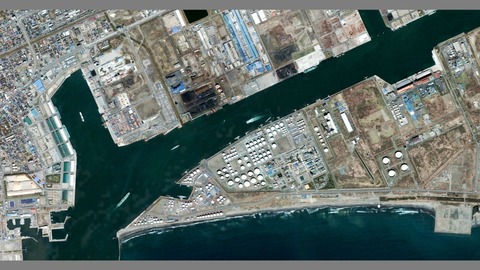} &
     \ig{60pt}{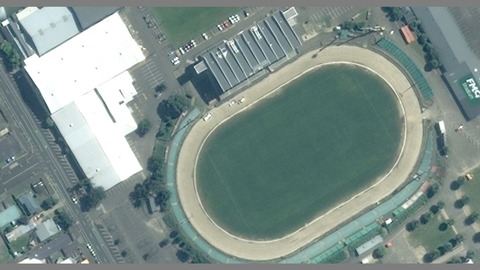} &
     \ig{60pt}{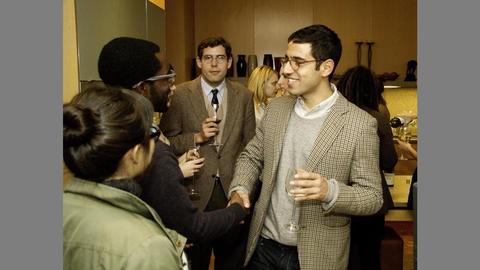} &
     \ig{60pt}{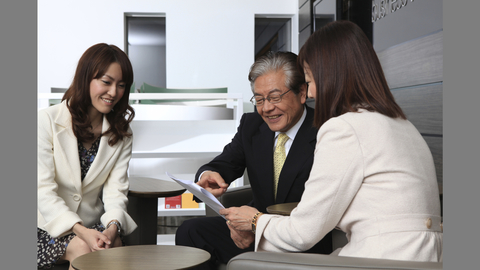} \\
     \end{tabular}}
   \caption{\textbf{Categories of the CAT2000 dataset}. }
\label{fig:cat2000}
\end{figure*}

Most existing saliency datasets only consider images which are roughly similar (e.g., natural images). However, the CAT2000 dataset has 20 diverse categories of images. While some of these categories have visual similarities (e.g., outdoor natural and outdoor man-made), other categories are quite distinct in appearance (e.g., satellite and cartoon). We assume that, in general, the categories are distinct enough to give rise to different saliency mechanisms. Figure~\ref{fig:cat2000} shows examples from the 20 classes. In our model $C_k$ corresponds to a particular $k$ category in the dataset.

During training the class of each image is known, but during testing we assume that it is unknown. Therefore a separate task of the network is to predict $p(C_k | \mathbf{I}, )$. This can be seen as a traditional classification problem, which neural networks excel at. We experimented using shared layers to compute both the saliency and the class, however we found that we achieve better performance with a separate set of layers to predict class. This separate network does not need to be so large because the classification problem is not very difficult and perfect classification performance is not necessary. Even in the case where classification is incorrect, if the network places emphasis on a similar class, the saliency prediction will still have good performance.

The gating network consists of 4 convolutional layers with max-pooling and 2 fully connected layers. The input to the gating network is down-sampled by 4. This is because the input image is of high resolution for saliency prediction, but such a high resolution is not necessary for category prediction. The parameters of the network can be found in Table~\ref{table:params}.

\begin{table}
  \centering
  \footnotesize
  \begin{tabular}{lll}
    \toprule
    & Layer & Hyper parameters\\
    \midrule
    \rt{VGG16}{18} & Input & 480 $\times$ 640 pixels \\
    & Conv1-1 & 64 (3 $\times$ 3) filters \\
    & Conv1-2 & 64 (3 $\times$ 3) filters \\
    & Pool1 & 2 $\times$ 2 max pooling \\
    & Conv2-1 & 128 (3 $\times$ 3) filters \\
    & Conv2-2 & 128 (3 $\times$ 3) filters \\
    & Pool2 & 2 $\times$ 2 max pooling \\
    & Conv3-1 & 256 (3 $\times$ 3) filters \\
    & Conv3-2 & 256 (3 $\times$ 3) filters \\
    & Conv3-3 & 256 (3 $\times$ 3) filters \\
    & Pool3 & 2 $\times$ 2 max pooling \\
    & Conv4-1 & 512 (3 $\times$ 3) filters \\
    & Conv4-2 & 512 (3 $\times$ 3) filters \\
    & Conv4-3 & 512 (3 $\times$ 3) filters \\
    & Pool4 & 2 $\times$ 2 max pooling (stride 1) \\
    & Conv5-1 & 512 (3 $\times$ 3) filters \\
    & Conv5-2 & 512 (3 $\times$ 3) filters \\
    & Conv5-3 & 512 (3 $\times$ 3) filters \\
    \midrule
    \rt{Experts}{2} & Conv-E-1 & 64 ($3 \times 3$) filters \\
    & Conv-E-2 & 16 ($1 \times 1$) filters \\
    & Center bias & $8 \times 6$ parameters \\
    \midrule
    \rt{Gating Network}{10} & Input & $120 \times 160$ pixels \\
    & Conv-G-1 & 32 (3 $\times$ 3) filters \\
    & Pool-G-1 & 2 $\times$ 2 max pooling \\
    & Conv-G-2 & 64 (3 $\times$ 3) filters \\
    & Pool-G-2 & 2 $\times$ 2 max pooling \\
    & Conv-G-3 & 128 (3 $\times$ 3) filters \\
    & Pool-G-3 & 2 $\times$ 2 max pooling \\
    & Conv-G-4 & 128 (3 $\times$ 3) filters \\
    & Pool-G-4 & 2 $\times$ 2 max pooling \\
    & Full1 & 128 units \\
    & Full2 & 20 units \\
    \bottomrule
  \end{tabular}
  \caption{Parameters of DNN.}
  \label{table:params}
  \end{table}

The output classification layer is typically normalized using the softmax function to obtain a probability prediction. Our goal is to use the softmax output to act as weights to choose which experts to use to compute saliency. Softmax outputs will favor a single expert, instead of a smooth mixture. To overcome this, we use a softmax function with a temperature parameter:

\begin{equation}
  p_\tau(C_k | \mathbf{I}) = \frac{\exp(\phi_k(\mathbf{I})/\tau)}{\sum_{i=1}^K\exp(\phi_i(\mathbf{I})/\tau)}
\end{equation}

\noindent where $\tau$ is the temperature parameter and $\phi_k$ is the output of the $k$th neuron in the last fully connected layer of the gating network. For higher temperatures, the distribution over classes will be more uniform, whereas for lower temperatures, the distribution will be sharper. In the extreme case where $\tau \to \infty$, the probability is equal for every class. For very high temperatures, the model simplifies to an ensemble of neural networks. For low temperatures, the model will choose a single path depending on the predicted class. In between these extremes, the model generates a weighted mixture. 


$p_\tau(C_k | \mathbf{I})$ gives a low dimensional representation of global scene characteristics. Previously it has been shown that outputs of the softmax at higher temperatures encode useful information~\cite{hinton-distillation}. Thus the outputs of the softmax are useful, even for an image that does not precisely fit into any of the predefined classes. While existing gist features~\cite{gist} encode global scene information, our model is more powerful because the scene representation is learned from training data simultaneously with the saliency prediction.
\vspace{-12pt}
\paragraph{Center bias}

The $p(s_i | x_i, C_k)$ term represents location bias in the eye tracking data. In most eye tracking experiments, the center bias effect causes fixations to appear closer to the center than the edges of the image~\cite{tatler-cb}. This effect is partially due to the photographer's bias that places objects in the center of images, and partially due to the framing effect experienced when viewing images on a computer monitor. We allow our network to learn the center bias as in previous works \cite{ml-net, deep-gaze1}, but also allow the center bias to vary between categories. It could be that some categories are more likely to be influenced by center bias. For example, the ``object'' category of the CAT2000 dataset is nearly always an object well placed in the center of the image.

To implement the location bias, we use the same formulation as \cite{ml-net}. The bias is modeled as a low resolution map of size $w_{cb} \times h_{cb}$. The map is upscaled and multiplied by the predicted saliency map to produce the center biased output. The elements of the map are all trainable parameters. The map is low resolution to limit overfitting. The initial values of the map are all set to $1$. In this initial condition, there is no location bias.

\subsection{Training}
\paragraph{Loss Function}
For training the network we utilize a loss function that considers both the accuracy of the resulting saliency map and the accuracy of the class prediction:
\begin{equation}
  \ell( \Theta; \mathbf{I}, \mathbf{x}, \mathbf{y}, \mathbf{t}) = \lambda_s \ell_s(\Theta; \mathbf{I}, \mathbf{x}, \mathbf{y}) + \lambda_c \ell_c(\Theta; \mathbf{I}, \mathbf{t}) 
  \label{eq:loss}
\end{equation}
\noindent where $\Theta$ represents the learnable parameters of the network. $\lambda_s$ and $\lambda_c$ are weights on the saliency loss ($\ell_s$) and the classification loss ($\ell_c$), respectively. $\mathbf{t}$ is the target category of the images used for training, and $\mathbf{y}$ is the ground-truth saliency map.

For the saliency loss ($\ell_s$) we use a similar cost function as \cite{ml-net}:
  \begin{multline}
\! \! \! \! \!    \ell_s(\Theta; \mathbf{I}, \mathbf{x}, \mathbf{y}) =
    \frac{1}{N}\sum_{i=1}^N \left \| \frac{p(s_i|\mathbf{I},x_i)}{\max_i p(s_i|\mathbf{I},x_i)} - y_i \right \|^2 / (\alpha-y_i)  \\
    + \frac{1}{K}\frac{1}{N}\sum_{k=1}^K\sum_{i=1}^N\lambda_{cb} \| 1 - p(s_i | x_i, C_k) \|^2
  \end{multline}

The first term normalizes the predicted saliency map and computes the squared error between the ground-truth ($y_i$) and the corresponding predicted saliency ($p(s_i|\mathbf{I},x_i)$). Provided that $\alpha >  \max{\mathbf{y}}$, the normalization by $1 / (\alpha-y_i)$ gives a larger weight to larger values in the ground-truth, and a smaller weight to smaller values. The second term is a regularization factor that encourages the network to learn saliency instead of only learning the center bias. $K$ is the number of categories, and $N$ is the number of pixels.

For the class loss ($\ell_c$) we use the standard categorical cross entropy:

\begin{equation}
\ell_c ( \Theta ; \mathbf{I}, \mathbf{t} ) = - \sum_k t_k \log ( p_{\tau=1}   ( C_k | \mathbf{I} ) )
\end{equation}
\noindent where $t_k$ is a binary variable that indicates if $k$ is the target class. For the class loss function we assume the softmax temperature $\tau$ is set to $1$.

\vspace{-12pt}
\paragraph{Training Procedure}
We begin with the weights from the pre-trained ML-net model \cite{ml-net}. This model was trained using the mouse-tracking data from the Salicon dataset \cite{salicon-db}. The mouse-tracking data was shown to be highly correlated with eye tracking data. We use these weights to initialize the convolutional layers of the VGG network. The weights of the last two layers for the expert networks, and the weights for the gating network are initialized using Glorot initialization \cite{glorot}. The center bias is initialized to be uniform.

We use the Adadelta algorithm for optimizing the parameters of the neural network \cite{adadelta}. We found that using the Adadelta algorithm achieves a higher accuracy than using stochastic gradient descent (as in \cite{ml-net}). Training for the classification loss and the saliency loss is done simultaneously using the loss function in Eq~\ref{eq:loss}. We stop the training if the validation loss does not decrease after 10 epochs. We use a batch size of 8 images.

During training we perform data augmentation by horizontally flipping a random 50\% of the samples in each batch. We conjecture that horizontal flipping does not significantly change the saliency (assuming ground-truth maps are also flipped), because much of the visual world is symmetric horizontally. We do not perform vertical flipping, as most of the stimuli are not symmetric vertically (e.g. grass and sky). Furthermore, there is an additional category in the CAT2000 dataset of vertically flipped images, so we do not wish to confuse the gating network on these samples.

\section{Performance Evaluation}
\label{sec:results}

\begin{figure*}
\setlength{\tabcolsep}{1pt}
\centering
   \begin{tabular}{ccccccc}
     \ig{60pt}{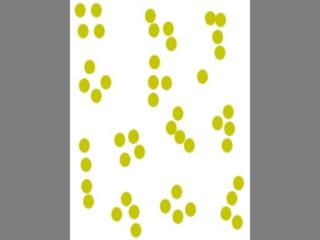} &
     \ig{60pt}{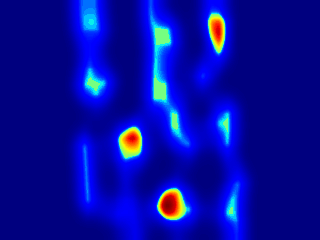} &
     \ig{60pt}{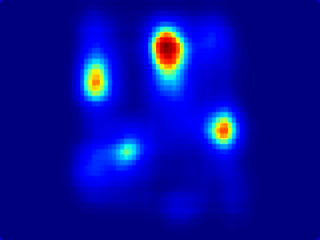} &
     \ig{60pt}{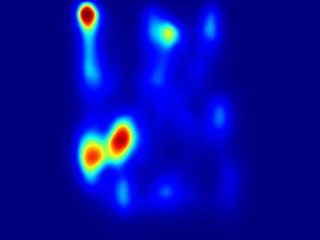} &
     \ig{60pt}{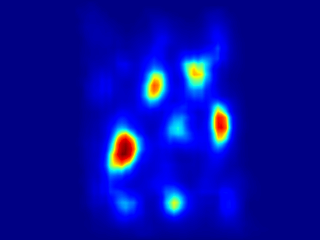} &
     \ig{60pt}{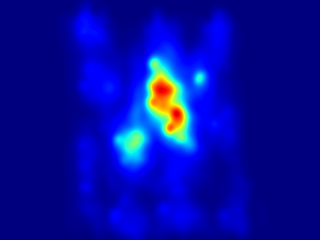} &
     \ig{60pt}{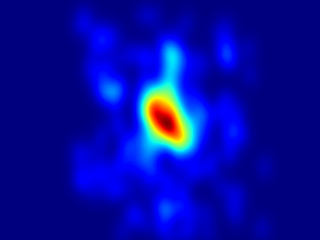} \\
     \ig{60pt}{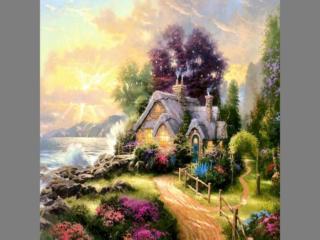} &
     \ig{60pt}{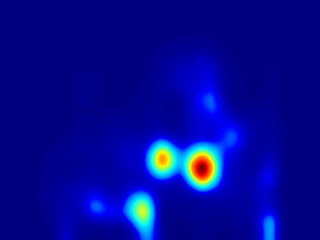} &
     \ig{60pt}{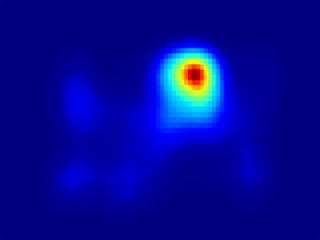} &
     \ig{60pt}{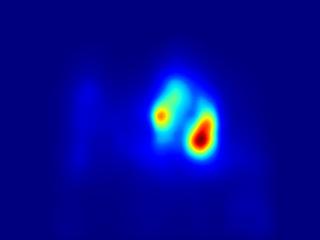} &
     \ig{60pt}{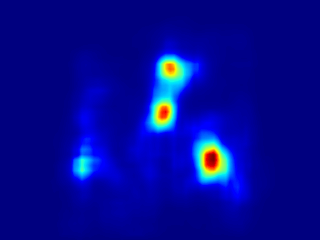} &
     \ig{60pt}{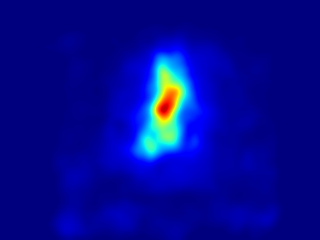} &
     \ig{60pt}{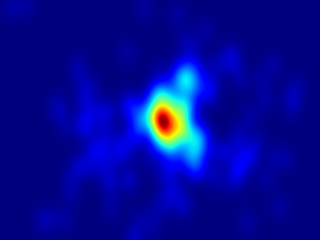} \\
     \ig{60pt}{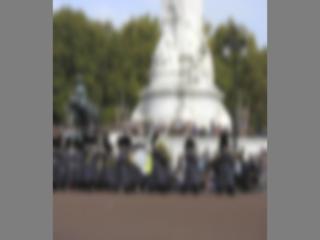} &
     \ig{60pt}{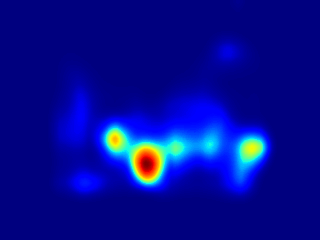} &
     \ig{60pt}{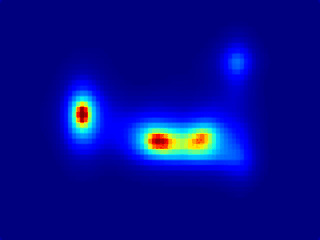} &
     \ig{60pt}{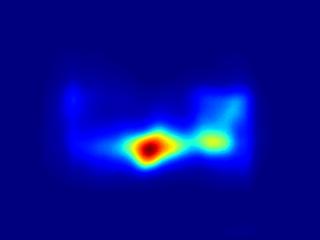} &
     \ig{60pt}{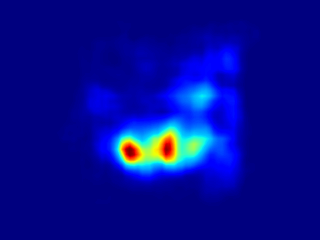} &
     \ig{60pt}{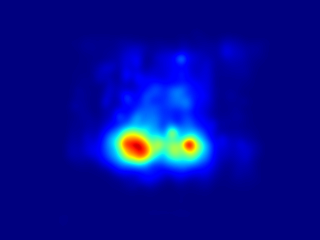} &
     \ig{60pt}{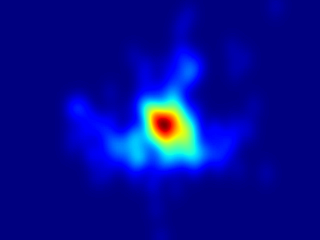} \\
     \ig{60pt}{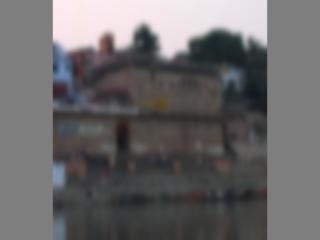} &
     \ig{60pt}{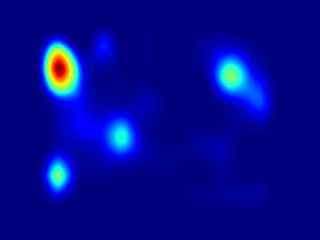} &
     \ig{60pt}{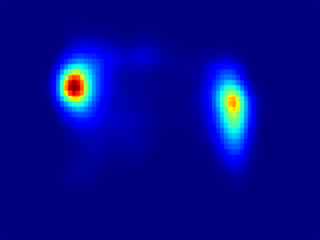} &
     \ig{60pt}{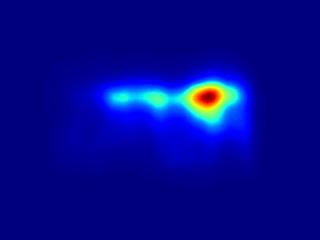} &
     \ig{60pt}{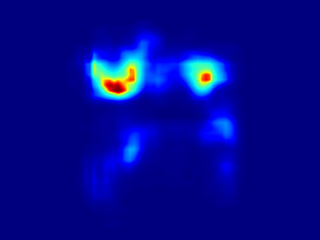} &
     \ig{60pt}{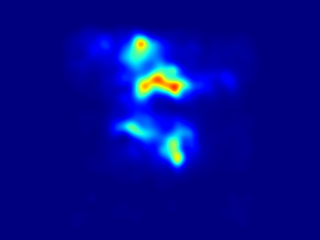} &
     \ig{60pt}{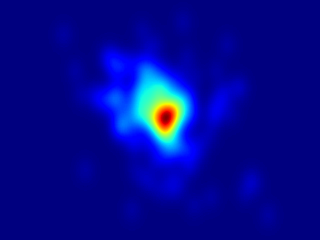} \\
     \ig{60pt}{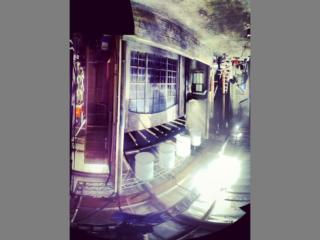} &
     \ig{60pt}{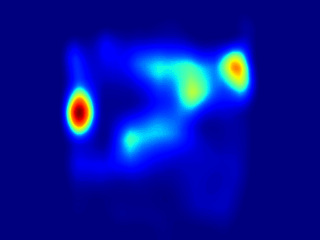} &
     \ig{60pt}{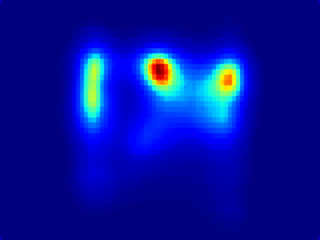} &
     \ig{60pt}{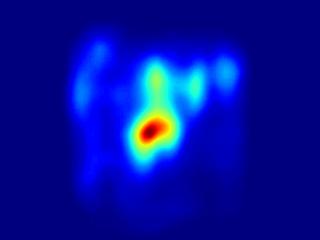} &
     \ig{60pt}{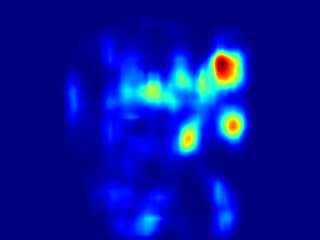} &
     \ig{60pt}{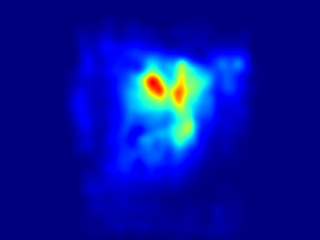} &
     \ig{60pt}{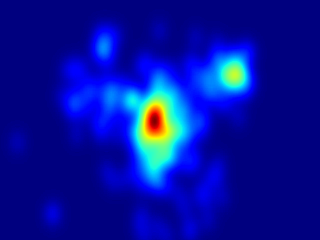} \\
     Stimuli & BMS \cite{bms} & Pan \etal \cite{junting} & Salicon \cite{salicon} & ML-Net \cite{ml-net} & \textbf{Ours} & Ground Truth \\
     \end{tabular}
   \caption{\textbf{Qualitative performance comparison}.}
\label{fig:qualitative}
\end{figure*}

\paragraph{Datasets}

We primarily use the CAT2000 dataset \cite{cat2000} because it comprises a sufficient number of images for deep learning (2000 in total), and it sorts the images into 20 different categories. Figure \ref{fig:cat2000} shows examples of the categories in the dataset. We use 5 fold cross-validation to train and test our model. We also submitted our model to the MIT CAT2000 saliency benchmark. This is a set of 2000 test images with the ground-truth held out, so that a model cannot over-fit the data.

We also test our model on the Salicon dataset \cite{salicon-db}. This is the largest available saliency dataset with 10,000 training images, 5,000 validation images, and 5,000 held out test images. Similar to the MIT benchmark, the test images do not have available ground-truth, but can be evaluated on a server. For this dataset, there is no ground-truth category information. We use the gating network trained using the CAT2000 dataset and set the learning rates of these layers to zero in training.

\vspace{-12pt}
\paragraph{Implementation Details}

We set $\lambda_c = 1$ and $\lambda_s = 10$. This is because the saliency loss is typically smaller than the classification loss, so these weights balance the two terms. $\lambda_{cb}$ is set to be $1$ and $\alpha$ is set to $1.1$ as in the original ML-Net model. $\tau$ is set to 10 to allow for smooth weights from the gating network. The code and the model will be released after publication to ensure reproducibility.


\vspace{-12pt}
\paragraph{Existing Models}
We have deliberately made our model similar to ML-Net \cite{ml-net} so that the advantage of our mixture based approach is clear. Our framework is general and can be applied to any base saliency model, however we chose to use ML-Net because of the public availability of the model and the code.

We compare with 3 deep neural network based models and one unsupervised model. We fine-tune ML-Net \cite{ml-net} on the same training and testing splits. We also compare with the Salicon model \cite{salicon}. This model is not publicly available, so we cannot fine-tune the results to the CAT2000 dataset. We are able to run the model on the CAT2000 images through the authors' website. Similarly, we run the pre-trained deep network from \cite{junting} without fine-tuning to the CAT2000 dataset. Finally, we compare with the BMS model \cite{bms}, as it is the best performing of the models that do not utilize training data.

\begin{figure}
  \centering
\ig{0.45\textwidth}{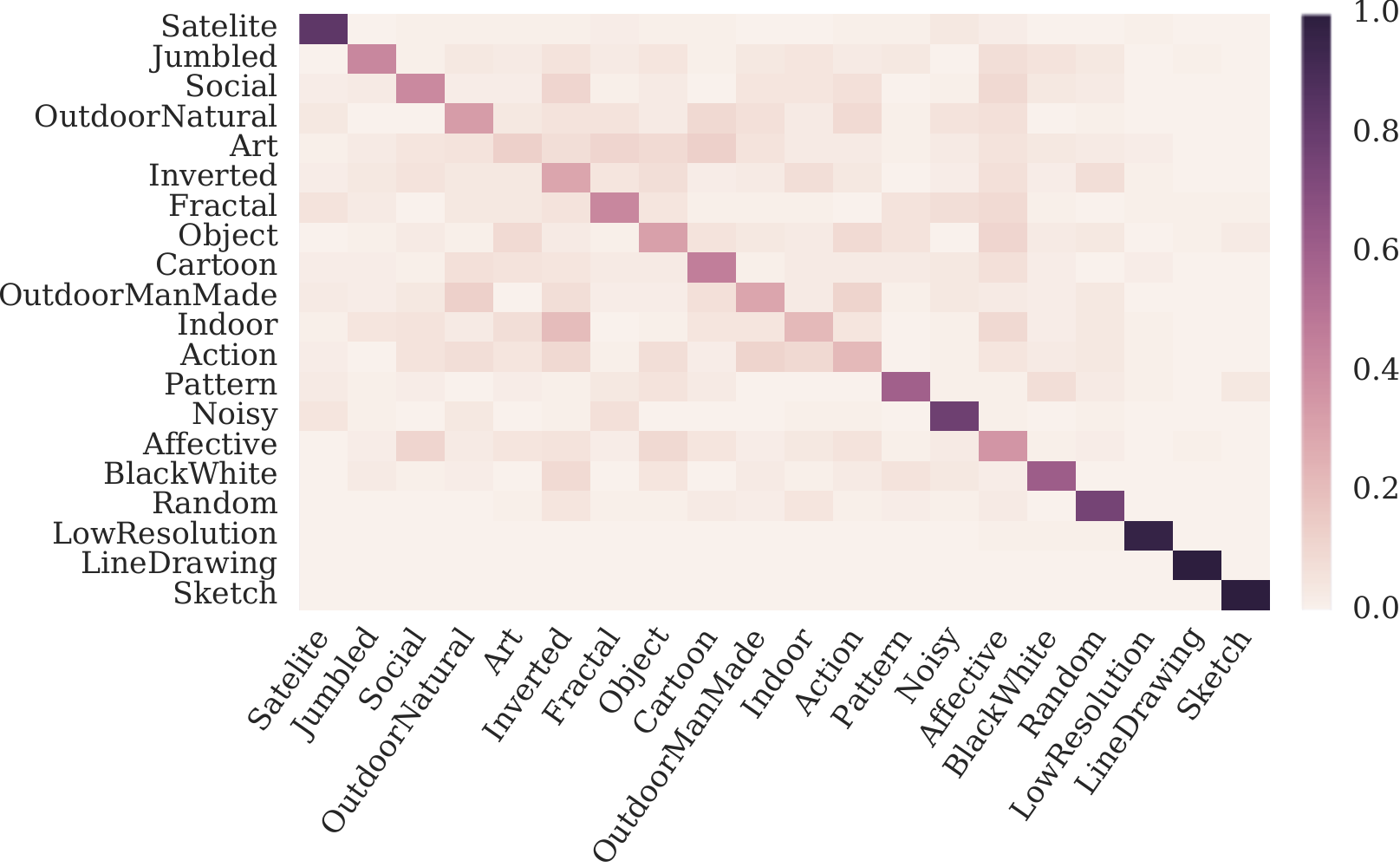}
   \caption{\textbf{Classification confusion matrix}. }
\label{fig:class}
\end{figure} 

\vspace{-12pt}
\paragraph{Evaluation Metrics}

Given the limitations of a single evaluation metric, it is best to evaluate using several metrics. We use the AUC-Borji, NSS, CC, and KLD metrics to perform our evaluation. We do not consider ``shuffled'' metrics such as SAUC, that attempt to correct for the center bias. Location bias is an important component of human visual attention, and so should be included in any model.

Table~\ref{tab:cat2000} shows the results of the 5-fold cross validation tests on the CAT2000 dataset. We achieve the best performance of the tested models. Table~\ref{tab:cat2000test} shows the performance on the held out test images. Our model achieves the second best performance behind the DeepFix model \cite{deepfix}. Compared with DeepFix our network is shallower with less parameters. The DeepFix model is not publicly available to test on the cross-validation data. Figure~\ref{fig:qualitative} shows example saliency maps generated by the tested models.

Table~\ref{tab:salicon} shows the results on the Salicon test set. Under this dataset our model does not have a performance advantage over ML-Net. This is largely because the dataset does not have diverse categories like the CAT-2000 dataset. Nevertheless, the performance of our model is roughly equivalent to the ML-net model.

\begin{table*}[!htb]
  \centering
\setlength{\tabcolsep}{2pt}
  \small
\begin{tabular}{lcccccccccccccccccccc}
  \toprule
 & \multicolumn{5}{c}{AUC} & \multicolumn{5}{c}{NSS} & \multicolumn{5}{c}{CC} & \multicolumn{5}{c}{KLD} \\
  \midrule
   & \textbf{Ours} & \cite{ml-net} & \cite{junting} & \cite{bms} & \cite{salicon} & \textbf{Ours} & \cite{ml-net} & \cite{junting} & \cite{bms} & \cite{salicon} & \textbf{Ours} & \cite{ml-net} & \cite{junting} & \cite{bms} & \cite{salicon} & \textbf{Ours} & \cite{ml-net} & \cite{junting} & \cite{bms} & \cite{salicon} \\
  \midrule
  All & \textbf{0.82} & 0.79 & 0.72 & 0.77 & 0.76 & \textbf{1.94} & 1.60 & 0.97 & 1.17 & 1.64 & \textbf{0.74} & 0.60 & 0.39 & 0.44 & 0.60 & \textbf{0.74} & 0.91 & 2.31 & 1.04 & 1.02 \\
  Action & 0.81 & 0.79 & \textbf{0.82} & 0.81 & 0.75 & \textbf{2.28} & 1.98 & 1.61 & 1.41 & 2.14 & \textbf{0.79} & 0.69 & 0.59 & 0.51 & 0.71 & \textbf{0.67} & 0.83 & 0.97 & 0.92 & 1.45 \\
  Affective & \textbf{0.82} & 0.80 & 0.81 & 0.81 & 0.78 & \textbf{2.42} & 2.19 & 1.65 & 1.46 & 2.27 & \textbf{0.80} & 0.72 & 0.56 & 0.51 & 0.73 & \textbf{0.65} & 0.80 & 1.28 & 0.95 & 1.17 \\
  Art & \textbf{0.81} & 0.77 & 0.79 & 0.77 & 0.76 & \textbf{1.80} & 1.46 & 1.30 & 1.16 & 1.53 & \textbf{0.72} & 0.57 & 0.52 & 0.46 & 0.59 & \textbf{0.71} & 0.91 & 0.89 & 0.95 & 1.05 \\
  BlackWhite & \textbf{0.83} & 0.80 & 0.54 & 0.78 & 0.78 & \textbf{2.26} & 1.86 & 0.14 & 1.23 & 2.00 & \textbf{0.75} & 0.62 & 0.05 & 0.42 & 0.65 & \textbf{0.78} & 0.94 & 4.90 & 1.08 & 0.98 \\
  Cartoon & \textbf{0.80} & 0.77 & 0.77 & 0.74 & 0.74 & \textbf{1.65} & 1.38 & 1.13 & 1.00 & 1.38 & \textbf{0.70} & 0.58 & 0.50 & 0.44 & 0.57 & \textbf{0.74} & 0.86 & 1.04 & 0.99 & 1.11 \\
  Fractal & \textbf{0.82} & 0.77 & 0.78 & 0.78 & 0.78 & \textbf{1.93} & 1.47 & 1.19 & 1.25 & 1.50 & \textbf{0.73} & 0.54 & 0.46 & 0.47 & 0.56 & \textbf{0.66} & 0.88 & 0.95 & 0.90 & 0.85 \\
  Indoor & \textbf{0.79} & 0.78 & 0.77 & 0.76 & 0.75 & \textbf{1.70} & 1.42 & 1.11 & 1.00 & 1.40 & \textbf{0.75} & 0.61 & 0.50 & 0.44 & 0.59 & \textbf{0.61} & 0.77 & 0.94 & 0.84 & 0.85 \\
  Inverted & \textbf{0.80} & 0.78 & 0.78 & 0.76 & 0.75 & \textbf{1.72} & 1.41 & 1.11 & 1.04 & 1.41 & \textbf{0.73} & 0.58 & 0.48 & 0.44 & 0.57 & \textbf{0.68} & 0.79 & 0.94 & 0.88 & 0.98 \\
  Jumbled & \textbf{0.78} & 0.76 & 0.76 & 0.75 & 0.72 & \textbf{1.48} & 1.24 & 1.00 & 0.92 & 1.17 & \textbf{0.72} & 0.60 & 0.50 & 0.46 & 0.56 & \textbf{0.66} & 0.76 & 0.73 & 0.75 & 0.73 \\
  LineDrawing & \textbf{0.82} & 0.79 & 0.40 & 0.76 & 0.74 & \textbf{1.79} & 1.40 & -0.39 & 1.06 & 1.26 & \textbf{0.73} & 0.57 & -0.17 & 0.43 & 0.50 & \textbf{0.67} & 0.85 & 4.69 & 1.15 & 1.11 \\
  LowResolution & \textbf{0.86} & 0.78 & 0.79 & 0.75 & 0.81 & \textbf{2.05} & 1.39 & 1.21 & 1.01 & 1.46 & \textbf{0.69} & 0.48 & 0.41 & 0.34 & 0.49 & \textbf{0.96} & 1.34 & 1.23 & 1.33 & 0.93 \\
  Noisy & \textbf{0.83} & 0.78 & 0.79 & 0.75 & 0.79 & \textbf{1.83} & 1.43 & 1.21 & 0.96 & 1.64 & \textbf{0.71} & 0.55 & 0.47 & 0.37 & 0.61 & \textbf{0.91} & 1.03 & 1.05 & 1.09 & 0.82 \\
  Object & \textbf{0.84} & 0.81 & 0.83 & 0.82 & 0.78 & \textbf{2.20} & 1.87 & 1.63 & 1.48 & 1.97 & \textbf{0.77} & 0.64 & 0.59 & 0.53 & 0.66 & \textbf{0.78} & 0.97 & 1.03 & 1.07 & 1.30 \\
  OutdoorManMade & \textbf{0.80} & 0.78 & 0.78 & 0.74 & 0.75 & \textbf{1.74} & 1.40 & 1.11 & 0.95 & 1.51 & \textbf{0.73} & 0.58 & 0.48 & 0.41 & 0.61 & \textbf{0.72} & 0.85 & 0.90 & 0.95 & 1.00 \\
  OutdoorNatural & \textbf{0.81} & 0.77 & 0.76 & 0.73 & 0.76 & \textbf{1.77} & 1.35 & 1.02 & 0.95 & 1.51 & \textbf{0.71} & 0.54 & 0.42 & 0.38 & 0.59 & \textbf{0.79} & 0.93 & 1.06 & 1.01 & 0.93 \\
  Pattern & \textbf{0.83} & 0.79 & 0.63 & 0.76 & 0.78 & \textbf{1.96} & 1.55 & 0.51 & 1.30 & 1.37 & \textbf{0.70} & 0.54 & 0.19 & 0.43 & 0.48 & \textbf{0.81} & 1.09 & 3.78 & 1.49 & 0.99 \\
  Random & \textbf{0.83} & 0.81 & 0.81 & 0.79 & 0.75 & \textbf{2.06} & 1.86 & 1.45 & 1.45 & 1.72 & \textbf{0.75} & 0.67 & 0.54 & 0.52 & 0.60 & \textbf{0.95} & 1.02 & 0.98 & 1.01 & 1.06 \\
  Satelite & \textbf{0.77} & 0.72 & 0.76 & 0.66 & 0.74 & \textbf{1.46} & 0.97 & 1.03 & 0.64 & 1.14 & \textbf{0.67} & 0.44 & 0.48 & 0.27 & 0.51 & \textbf{0.77} & 0.92 & 0.91 & 1.04 & 0.79 \\
  Sketch & \textbf{0.88} & 0.86 & 0.27 & 0.87 & 0.81 & \textbf{2.64} & 2.39 & -0.97 & 2.11 & 2.52 & \textbf{0.80} & 0.71 & -0.32 & 0.63 & 0.74 & \textbf{0.75} & 0.90 & 17.13 & 1.50 & 1.31 \\
  Social & \textbf{0.80} & 0.79 & 0.80 & 0.76 & 0.77 & \textbf{2.04} & 1.89 & 1.31 & 0.97 & 1.91 & \textbf{0.79} & 0.73 & 0.54 & 0.41 & 0.72 & \textbf{0.58} & 0.70 & 0.91 & 0.96 & 1.01 \\
  \bottomrule
  \end{tabular}
\caption{\textbf{CAT2000 cross validation set results.}}
\label{tab:cat2000}
\end{table*}

\begin{table*}[!htb]
  \small
  \centering
\begin{tabular}{lcccccccc}
  \toprule
Model  & AUC-Judd & SIM & EMD & AUC-Borji & sAUC & CC & NSS & KL \\
  \midrule
Baseline: infinite humans & 0.90 & 1 & 0 & 0.84 & 0.62 & 1 & 2.85 & 0 \\
DeepFix \cite{deepfix} & 0.87 & 0.74 & 1.15 & 0.81 & 0.58 & 0.87 & 2.28 & 0.37 \\
Our Model & 0.86 & 0.66 & 1.63 & 0.82 & 0.58 & 0.76 & 1.92 & 0.62 \\
iSEEL\cite{iseel} & 0.84 & 0.62 & 1.78 & 0.81 & 0.59 & 0.66 & 1.67 & 0.92 \\
Ensembles of Deep Networks (eDN) \cite{edn} & 0.85 & 0.52 & 2.64 & 0.84 & 0.55 & 0.54 & 1.30 & 0.97 \\
Boolean Map based Saliency (BMS) \cite{bms} & 0.85 & 0.61 & 1.95 & 0.84 & 0.59 & 0.67 & 1.67 & 0.83 \\
Judd Model\cite{judd} & 0.84 & 0.46 & 3.60 & 0.84 & 0.56 & 0.54 & 1.30 & 0.94 \\
  \bottomrule
  \end{tabular}
\caption{\textbf{CAT2000 test set results.}}
\label{tab:cat2000test}
\end{table*}

\begin{table}[tb]
  \small
  \label{table:salicon}
  \centering
  \begin{tabular}{lccc}
    \toprule
    & CC & SAUC & AUC Judd\\ 
    \midrule
    \textbf{Our Model} & 0.730  & 0.771 & 0.861 \\
    ML-Net \cite{ml-net} & 0.7430 & 0.7680 & 0.8660 \\
    Deep Convnet \cite{junting}& 0.6220& 0.7240& 0.8580\\
    Shallow Convnet \cite{junting}& 0.5957& 0.6698& 0.8364\\
    Rare 2012 Improved~\cite{rare2012}& 0.5108& 0.6644& 0.8148\\
    Baseline: BMS~\cite{bms}& 0.4268& 0.6935& 0.7899\\
    Baseline: GBVS~\cite{gbvs}& 0.4212& 0.6303& 0.7899\\
    Baseline: Itti~\cite{itti}& 0.2046& 0.6101& 0.6669\\
    \bottomrule
  \end{tabular}
  \caption{\textbf{Salicon test set results.}}
\label{tab:salicon}
  \end{table}

\vspace{-12pt}
\paragraph{Comparison with averaging ensemble}
Given that the mixture model can be seen as an ensemble, it is important to compare the performance with a vanilla ensemble. We generate an ensemble of 5 networks where each network is trained on a random 80\% of the training split. The saliency maps from each ensemble member are averaged to produce the final saliency map. Our model still achieves superior performance compared with the ensemble (Table~\ref{tab:ensemble}). Additionally, our model is significantly faster and requires less storage space due to the weight sharing.

\vspace{-12pt}
\paragraph{Classification performance}
Figure~\ref{fig:class} shows the confusion matrix of the classification output of the gating network. Some classes of images are very distinct and easy to classify (e.g., satellite images and line drawings), while other categories lack distinctive characteristics for good classification performance (e.g., action images and art images). For the easily classified images our model will favor a single expert. This is appropriate because these images are very distinct. For less easily classified images, our model will favor a smoother mixture of related experts.

\section{Conclusions}

We propose a novel mixture of experts based model to predict image saliency. Our model uses global scene information in addition to local information from a convolutional neural network. The global scene information is trained in a supervised fashion from the diverse categories of the CAT2000 eye-tracking dataset. The model is used to create a final saliency map as a mixture of saliency maps predicted by networks that are experts for a particular class. Our model improves results compared with a similar model that does not use global scene context, and achieves better performance than several other deep-learning based models.

\begin{table}
  \small
  \centering
  \begin{tabular}{lllll}
    \toprule
    Method & AUC & NSS & CC & KLD \\
    \midrule
    Averaging Ensemble & 0.74 & 1.47 & 0.54 & 0.93 \\
    Our Model & 0.82 & 1.91 & 0.74 & 0.78 \\
    \bottomrule
  \end{tabular}
  \caption{\textbf{Ensemble results on CAT2000 cross validation set.} Results are shown only for the first split of the data.}
  \label{tab:ensemble}
  \end{table}


Our model represents an important step forward because it can adapt to vastly different types of inputs. This is important because a saliency model should not be limited to similar images. Saliency may be needed in diverse application domains such as driving assistance systems or to predict web behavior. One could build separate models for separate applications, but it is more useful to have a unified model. In this paper our model is limited to the stimuli types in the CAT2000 database, but our mixture of experts formulation is capable of modeling more varied stimuli.


\textbf{Acknowledgments}
We thank nVidia Corporation for the donation of the Titan X GPU used in this work.

{\small
\bibliographystyle{ieee}
\bibliography{refs}

\begin{thebibliography}{10}\itemsep=-1pt

\bibitem{borji-boost}
A.~Borji.
\newblock Boosting bottom-up and top-down visual features for saliency
  estimation.
\newblock In {\em CVPR}, 2012.

\bibitem{borji-scene}
A.~Borji and L.~Itti.
\newblock Scene classification with a sparse set of salient regions.
\newblock In {\em IEEE International Conference on Robotics and Automation
  (ICRA)}, 2011.

\bibitem{borji-survey}
A.~Borji and L.~Itti.
\newblock State-of-the-art in visual attention modeling.
\newblock {\em TPAMI}, 2013.

\bibitem{cat2000}
A.~Borji and L.~Itti.
\newblock Cat2000: A large scale fixation dataset for boosting saliency
  research.
\newblock {\em CVPR 2015 workshop on "Future of Datasets"}, 2015.
\newblock arXiv preprint arXiv:1505.03581.

\bibitem{henderson-real}
J.~R. Brockmole and J.~M. Henderson.
\newblock Using real-world scenes as contextual cues for search.
\newblock {\em Visual Cognition}, 2006.

\bibitem{aim}
N.~Bruce and J.~Tsotsos.
\newblock Saliency, attention, and visual search: An information theoretic
  approach.
\newblock {\em Journal of Vision}, 2009.

\bibitem{chun}
M.~M. Chun and Y.~Jiang.
\newblock Contextual cueing: Implicit learning and memory of visual context
  guides spatial attention.
\newblock {\em Cognitive psychology}, 1998.

\bibitem{bot-top}
C.~E. Connor, H.~E. Egeth, and S.~Yantis.
\newblock Visual attention: bottom-up versus top-down.
\newblock {\em Current Biology}, 2004.

\bibitem{ml-net}
M.~Cornia, L.~Baraldi, G.~Serra, and R.~Cucchiara.
\newblock A deep multi-level network for saliency prediction.
\newblock In {\em International Conference on Pattern Recognition (ICPR)},
  2016.

\bibitem{frintrop-slam}
S.~Frintrop and P.~Jensfelt.
\newblock Attentional landmarks and active gaze control for visual slam.
\newblock {\em IEEE Transactions on Robotics}, 2008.

\bibitem{glorot}
X.~Glorot and Y.~Bengio.
\newblock Understanding the difficulty of training deep feedforward neural
  networks.
\newblock In {\em International Conference on Artificial Intelligence and
  Statistics}, 2010.

\bibitem{gbvs}
J.~Harel, C.~Koch, and P.~Perona.
\newblock Graph-based visual saliency.
\newblock In {\em NIPS}, 2007.

\bibitem{hinton-distillation}
G.~Hinton, O.~Vinyals, and J.~Dean.
\newblock Distilling the knowledge in a neural network.
\newblock {\em arXiv preprint arXiv:1503.02531}, 2015.

\bibitem{HouZhang}
X.~Hou and L.~Zhang.
\newblock Saliency detection: A spectral residual approach.
\newblock In {\em CVPR}, 2007.

\bibitem{salicon}
X.~Huang, C.~Shen, X.~Boix, and Q.~Zhao.
\newblock Salicon: Reducing the semantic gap in saliency prediction by adapting
  deep neural networks.
\newblock {\em ICCV}, 2015.

\bibitem{itti}
L.~Itti, C.~Koch, and E.~Niebur.
\newblock A model of saliency-based visual attention for rapid scene analysis.
\newblock {\em TPAMI}, 1998.

\bibitem{hinton-mixture}
R.~Jacobs, S.~Jordan, Michaeland~Nowlan, and G.~Hinton.
\newblock Adaptive mixtures of local experts.
\newblock {\em Neural computation}, 1991.

\bibitem{vig}
S.~Jetley, N.~Murray, and E.~Vig.
\newblock End-to-end saliency mapping via probability distribution prediction.
\newblock {\em CVPR}, 2016.

\bibitem{salicon-db}
M.~Jiang, S.~Huang, J.~Duan, and Q.~Zhao.
\newblock Salicon: Saliency in context.
\newblock In {\em CVPR}, 2015.

\bibitem{judd}
T.~Judd, K.~Ehinger, F.~Durand, and A.~Torralba.
\newblock Learning to predict where humans look.
\newblock In {\em ICCV}, 2009.

\bibitem{kienzle}
W.~Kienzle, F.~A. Wichmann, B.~S. {\`I}lkopf, and M.~O. Franz.
\newblock A nonparametric approach to bottom-up visual saliency.
\newblock In {\em NIPS}, 2007.

\bibitem{deepfix}
S.~S. Kruthiventi, K.~Ayush, and R.~V. Babu.
\newblock Deepfix: A fully convolutional neural network for predicting human
  eye fixations.
\newblock {\em arXiv preprint arXiv:1510.02927}, 2015.

\bibitem{deep-gaze1}
M.~K{\"u}mmerer, L.~Theis, and M.~Bethge.
\newblock Deep gaze i: Boosting saliency prediction with feature maps trained
  on imagenet.
\newblock {\em arXiv preprint arXiv:1411.1045}, 2014.

\bibitem{deep-gaze2}
M.~K{\"u}mmerer, T.~S. Wallis, and M.~Bethge.
\newblock Deepgaze ii: Reading fixations from deep features trained on object
  recognition.
\newblock {\em arXiv preprint arXiv:1610.01563}, 2016.

\bibitem{kunar}
M.~A. Kunar, S.~J. Flusberg, and J.~M. Wolfe.
\newblock Contextual cuing by global features.
\newblock {\em Perception \& psychophysics}, 2006.

\bibitem{tree-net}
S.~Lee, S.~Purushwalkam, M.~Cogswell, D.~Crandall, and D.~Batra.
\newblock Why m heads are better than one: Training a diverse ensemble of deep
  networks.
\newblock {\em arXiv preprint arXiv:1511.06314}, 2015.

\bibitem{lenc}
K.~Lenc and A.~Vedaldi.
\newblock Understanding image representations by measuring their equivariance
  and equivalence.
\newblock In {\em CVPR}, 2015.

\bibitem{gist}
A.~Oliva and A.~Torralba.
\newblock Modeling the shape of the scene: A holistic representation of the
  spatial envelope.
\newblock {\em International Journal of Computer Vision}, 2001.

\bibitem{junting}
J.~Pan, E.~Sayrol, X.~{Giro-i-Nieto}, K.~McGuinness, and N.~E. O'Connor.
\newblock Shallow and deep convolutional networks for saliency prediction.
\newblock {\em CVPR}, 2016.

\bibitem{itti-peters}
R.~Peters and L.~Itti.
\newblock Beyond bottom-up: Incorporating task-dependent influences into a
  computational model of spatial attention.
\newblock {\em CVPR}, 2007.

\bibitem{rare2012}
N.~Riche, M.~Mancas, M.~Duvinage, M.~Mibulumukini, B.~Gosselin, and T.~Dutoit.
\newblock Rare2012: A multi-scale rarity-based saliency detection with its
  comparative statistical analysis.
\newblock {\em Signal Processing: Image Communication}, 2013.

\bibitem{vgg}
K.~Simonyan and A.~Zisserman.
\newblock Very deep convolutional networks for large-scale image recognition.
\newblock {\em International Conference on Learning Representations}, 2014.

\bibitem{inception}
C.~Szegedy, W.~Liu, Y.~Jia, P.~Sermanet, S.~Reed, D.~Anguelov, D.~Erhan,
  V.~Vanhoucke, and A.~Rabinovich.
\newblock Going deeper with convolutions.
\newblock In {\em CVPR}, 2015.

\bibitem{tatler-cb}
B.~W. Tatler.
\newblock The central fixation bias in scene viewing: Selecting an optimal
  viewing position independently of motor biases and image feature
  distributions.
\newblock {\em Journal of Vision}, 2007.

\bibitem{iseel}
H.~R. Tavakoli, A.~Borji, J.~Laaksonen, and E.~Rahtu.
\newblock Exploiting inter-image similarity and ensemble of extreme learners
  for fixation prediction using deep features.
\newblock {\em arXiv preprint arXiv:1610.06449}, 2016.

\bibitem{torralba2006}
A.~Torralba, A.~Oliva, M.~S. Castelhano, and J.~M. Henderson.
\newblock Contextual guidance of eye movements and attention in real-world
  scenes: the role of global features in object search.
\newblock {\em Psychological Review}, 2006.

\bibitem{edn}
E.~Vig, M.~Dorr, and D.~Cox.
\newblock Large-scale optimization of hierarchical features for saliency
  prediction in natural images.
\newblock In {\em CVPR}, 2014.

\bibitem{walther}
D.~Walther, U.~Rutishauser, C.~Koch, and P.~Perona.
\newblock Selective visual attention enables learning and recognition of
  multiple objects in cluttered scenes.
\newblock {\em Computer Vision and Image Understanding}, Oct. 2005.

\bibitem{yarbus}
A.~L. Yarbus.
\newblock {\em Eye Movements and Vision}.
\newblock 1967.

\bibitem{adadelta}
M.~D. Zeiler.
\newblock Adadelta: an adaptive learning rate method.
\newblock {\em arXiv preprint arXiv:1212.5701}, 2012.

\bibitem{bms}
J.~Zhang and S.~Sclaroff.
\newblock Saliency detection: A boolean map approach.
\newblock In {\em ICCV}, 2013.

\bibitem{sun}
L.~Zhang, M.~H. Tong, T.~K. Marks, H.~Shan, and G.~W. Cottrell.
\newblock Sun: A bayesian framework for saliency using natural statistics.
\newblock {\em Journal of Vision}, 2008.

\end{thebibliography}
}

\end{document}